\documentclass[11pt,a4paper]{article}
\usepackage[utf8]{inputenc}
\usepackage[hyperref]{naaclhlt2018}
\usepackage{times}
\usepackage{latexsym}
\usepackage{graphicx}
\usepackage{subfigure}
\usepackage{float}
\usepackage{enumerate}
\usepackage[american]{babel}
\usepackage{cleveref}

\usepackage{xcolor}

\crefname{section}{§}{§§}
\Crefname{section}{§}{§§}

\usepackage{url}

\aclfinalcopy 
\def\aclpaperid{260} 

\title{Comparatives, Quantifiers, Proportions:\\A Multi-Task Model for the Learning of Quantities from Vision}

\author{Sandro Pezzelle$^*$, Ionut-Teodor Sorodoc$^*$$^\dagger$, Raffaella Bernardi$^*$$^\ddagger$\\
	      $^*$CIMeC - Center for Mind/Brain Sciences, University of Trento\\
	      $^\dagger$Universitat Pompeu Fabra Barcelona\\
	      $^\ddagger$DISI - Department of Information Engineering and Computer Science, University of Trento\\
	      $^*${\tt sandro.pezzelle@unitn.it}, $^\dagger${\tt ionutteodor.sorodoc@upf.edu}\\
	      $^\ddagger${\tt raffaella.bernardi@unitn.it}}

\date{}

\begin{document}
\maketitle
\begin{abstract}
The present work investigates whether different quantification mechanisms (set comparison, vague quantification, and proportional estimation) can be jointly learned from visual scenes by a multi-task computational model. The motivation is that, in humans, these processes underlie the same cognitive, non-symbolic ability, which allows an automatic estimation and comparison of set magnitudes. We show that when information about lower-complexity tasks is available, the higher-level proportional task becomes more accurate than when performed in isolation. Moreover, the multi-task model is able to generalize to unseen combinations of target/non-target objects. Consistently with behavioral evidence showing the interference of absolute number in the proportional task, the multi-task model no longer works when asked to provide the number of target objects in the scene.


\end{abstract}



\section{Introduction}
\label{sec:introduction}

Understanding and producing sentences like `There are \textit{more} cars than parking lots', `\textit{Most} of the supporters wear blue t-shirts', `\textit{Twenty percent} of the trees have been planted last year', or `\textit{Seven} students passed the exam', is a fundamental competence which allows speakers to communicate information about quantities. Crucially, the type of information conveyed by these expressions, as well as their underlying cognitive mechanisms, are not equivalent, as suggested by evidence from linguistics, language acquisition, and cognition.

First, comparatives (`more', `less'), quantifiers (`some', `most', `all'), and proportions (`$20$\%', `two thirds') express a comparison or relation \textit{between sets} (e.g., between the set of cars and the set of parking lots). Such relational information is rather coarse when expressed by comparatives and vague quantifiers, more precise when denoted by proportions. In contrast, numbers (`one', `six', `twenty-two') denote the exact, absolute cardinality of the items belonging to \textit{one set} (e.g., the set of students who passed the exam).

Second, during language acquisition, these expressions are neither learned at the same time nor governed by the same rules. Recent evidence showed that children can understand comparatives at around $3$.$3$ years~\cite{odic2013,bryant2017}, with quantifiers being learned a few months later, at around $3$.$4$-$3$.$6$ years~\cite{hurewitz2006,minai2006,halberda2008}. Crucially, knowing the meaning of numbers, an ability that starts not before the age of $3$.$5$ years~\cite{le2007}, is not required to understand and use these expressions. As for proportions, they are acquired significantly later, being fully mastered only at the age of $9$ or $10$~\cite{hartnett1998,moss1999,sophian2000}.

Third, converging evidence from cognition and neuroscience supports the hypothesis that some important components of these expressions of quantity are grounded on a preverbal, non-symbolic system representing magnitudes~\cite{piazza2010}. This system, often referred to as Approximate Number System ({ANS}), is invariant to the sensory modality and almost universal in the animal domain, and consists in the ability of holistically extracting and comparing approximate numerosities~\cite{piazza2016}. In humans, it is present since the youngest age, with $6$-month-old infants being able to automatically compare sets and combine them by means of proto-arithmetical operations~\cite{xu2000,mccrink2004}. Since it obeys Weber's law, according to which highly differing sets (e.g. $2$:$8$) are easier to discriminate than highly similar sets (e.g. $7$:$8$), {ANS} has been recently claimed to be a \textit{ratio-based} mechanism~\cite{sidney2017,matthews2016ind}. In support of this, behavioral findings indicate that, in non-symbolic contexts (e.g. visual scenes), proportional values are extracted holistically, i.e. without relying on the pre-computed cardinalities of the sets~\cite{fabbri2012,yang2015}. Indeed, people are fairly accurate in providing the proportion of targets in a scene, even in high-speed settings~\cite{healey1996,treisman2006}. Similarly, in briefly-presented scenes, the interpretation of quantifiers is shown to be best described by proportional information~\cite{anonymous}.

Altogether, this suggests that performing ($1$) set comparison, ($2$) vague quantification, and ($3$) proportional estimation, which all rely on information regarding relations among sets, underlies increasingly-complex steps of the same mechanism. Notably, such complexity would range from `more/less' judgements to proportional estimation, as suggested by the increasing precision of {ANS} through years~\cite{halberda2008b}, the reported boundary role of `half' in early proportional reasoning~\cite{spinillo1991}, and the different age of acquisition of the corresponding linguistic expressions. Finally, the ratio-based operation underlying these task would be different from (and possibly conflicting with) that of estimating the absolute numerosity of one set. Indeed, absolute numbers are found to interfere with the access to proportions~\cite{fabbri2012}.

\begin{figure}[t!]
\includegraphics[width=\columnwidth]{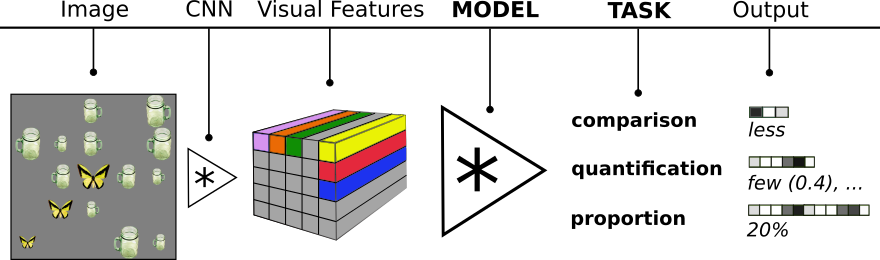}
\caption{Toy representation of the quantification tasks and corresponding outputs explored in the paper. Note that quantification always refers to animals (target set).}\label{fig:toymodel}
\end{figure}

Inspired by this converging evidence, the present work proposes a computational framework to explore various quantification tasks in the visual domain (see Figure~\ref{fig:toymodel}). In particular, we investigate whether ratio-based quantification tasks can be modeled by a single, multi-task learning neural network. Given a synthetic scene depicting animals (in our setting, the `target' objects) and artifacts (`non-target'), our model is designed to jointly perform all the tasks by means of an architecture that reflects their increasing complexity.\footnote{The dataset and the code can be downloaded from \texttt{github.com/sandropezzelle/multitask-quant}} To perform proportional estimation (the most complex), the model builds on the representations learned to perform vague quantification and, in turn, set comparison (the least complex). We show that the multi-task model achieves both higher accuracy and higher generalization power compared to the one-task models. In contrast, we prove that introducing the absolute number task in the loop is not beneficial and indeed hurts the performance.

Our main contribution lies in the novel application and evaluation of a multi-task learning architecture on the task of jointly modeling 3 different quantification operations. On the one hand, our results confirm the interdependency of the mechanisms underlying the tasks of set comparison, vague quantification, and proportional estimation. On the other, we provide further evidence on the effectiveness of these computational architectures.

\section{Related Work}
\label{sec:related_work}

\subsection{Quantities in Language \& Vision}

In recent years, the task of extracting quantity information from visual scenes has been tackled via Visual Question Answering ({VQA}). Given a real image and a natural language question, a {VQA} computational model is asked to understand the image, the linguistic query, and their interaction to provide the correct answer. So-called \textit{count} questions, i.e. `How many \textit{Xs} have the property \textit{Y}?', are very frequent and have been shown to be particularly challenging for any model~\cite{antol2015,malinowski2015,ren2015,fukui2016}. The difficulty of the task has been further confirmed by the similarly poor performance achieved even on the `diagnostic' datasets, which include synthetic visual scenes depicting geometric shapes~\cite{clevr2017,suhr2017}.

Using Convolutional Neural Networks ({CNN}), a number of works in Computer Vision ({CV}) have proposed specific architectures for counting digits~\cite{segui2015}, people in the crowd~\cite{zhang2015}, and penguins~\cite{arteta2016}. With a more cognitive flavor,~\citet{chatto2016} employed a `divide-and-conquer' strategy to split the image into subparts and count the objects in each subpart by mimicking the `subitizing' mechanism (i.e. numerosities up to $3$-$4$ can be rapidly and accurately appreciated). Inspired by the same cognitive ability is~\citet{sos2015}, who trained a {CNN} to detect and count the salient objects in the image. Except~\citet{suhr2017}, who evaluated models against various types of quantity expressions (including existential quantifiers), these works were just focused on the absolute number.

More akin to our work is~\citet{stoianov2012}, who showed that hierarchical generative models learn {ANS} as a statistical property of (synthetic) images. Their networks were tested on the task of set comparison (`more/less') and obtained 93\% accuracy. A few studies specifically focused on the learning of quantifiers.~\citet{sorodoc2016} proposed a model to assign the correct quantifier to synthetic scenes of colored dots, whereas~\citet{sorodoc2018} operationalized the same task in a {VQA} fashion, using real images and object-property queries (e.g. `How many \textit{dogs} are \textit{black}?'). Overall, the results of these studies showed that vague quantification can be learned by neural networks, though the performance is much lower when using real images and complex queries. Finally,~\citet{pezzelle2017} investigated the difference between the learning of cardinals and quantifiers from visual scenes, showing that they require two distinct computational operations. To our knowledge, this is the first attempt to jointly investigate the whole range of quantification mechanisms. Moreover, we are the first to exploit a multi-task learning paradigm for exploring the interactions between set comparison, vague quantification, and proportions.

\subsection{Multi-Task Learning}\label{sec:multi}

Multi-Task Learning ({MTL}) has been shown to be very effective for a wide range of applications in machine learning (for an overview, see~\citet{ruder2017}). The core idea is that different and yet related tasks can be jointly learned by a multi-purpose model rather than by separate and highly fine-tuned models. Since they share representations between related (or `auxiliary') tasks, multi-task models are more robust and generalize better than single-task models. Successful applications of {MTL} have been proposed in {CV} to improve object classification~\cite{girshick2015}, face detection and rotation~\cite{zhang2014,yim2015}, and to jointly perform a number of tasks as object detection, semantic segmentation, etc.~\cite{misra2016,li2016}. Though, recently, a few studies applied {MTL} techniques to either count or estimate the number of objects in a scene~\cite{sun2017,sindagi2017}, to our knowledge none of them were devoted to the learning of various quantification mechanisms.

In the field of natural language processing ({NLP}), {MTL} turned out to be beneficial for machine translation~\cite{luong2016} and for a range of tasks such as chunking, tagging, semantic role labelling, etc.~\cite{collobert2011,sogaard2016,bingel2017}. In particular,~\citet{sogaard2016} showed the benefits of keeping low-level tasks at the lower layers of the network, a setting which enables higher-level tasks to make a better use of the shared representations. Since this finding was also in line with previous evidence suggesting a natural order among different tasks~\cite{shen2005}, further work proposed {MTL} models in which several increasingly-complex tasks are hierarchically ordered~\cite{hashimoto2017}. The intuition behind this architecture, referred to as `joint many-task model' in the source paper~\cite{hashimoto2017}, as well as its technical implementation, constitute the building blocks of the model proposed in the present study.

\section{Tasks and Dataset}
\label{sec:dataset}

\subsection{Tasks}\label{sec:tasks}

Given a visual scene depicting a number of animals (targets) and artifacts (non-targets), we explore the following tasks, represented in Figure~\ref{fig:toymodel}:
\begin{enumerate}[(a)]
\item set comparison (hence, \textbf{{setComp}}), i.e. judging whether the targets are `more', `same', `less' than non-targets;
\item vague quantification (hence, \textbf{{vagueQ}}), i.e. predicting the probability to use each of the $9$ quantifiers (`none', `almost none', `few', `the smaller part', `some', `many', `most', `almost all', `all') to refer to the target set;
\item proportional estimation (hence, \textbf{{propTarg}}), i.e. predicting the proportion of targets choosing among $17$ ratios, ranging from $0$ to $100$\%.
\end{enumerate}

Tasks (a) and (c) are operationalized as classification problems and evaluated through accuracy. That is, only one answer out of $3$ and $17$, respectively, is considered as correct. Given the vague status of quantifiers, whose meanings are `fuzzy' and overlapping, task (b) is evaluated by means of Pearson's correlation (\textit{r}) between the predicted and the ground-truth probability vector (cf.~\cref{sec:datap}), for each datapoint.\footnote{We also experimented with Mean Average Error and dot product and found the same patterns of results (not reported).} The overall \textit{r} is obtained by averaging these scores. It is worth mentioning that we could either evaluate (b) in terms of a classification task or operationalize (a) and (c) in terms of a correlation with human responses. The former evaluation is straightforward and can be easily carried out by picking the quantifier with the highest probability. The latter, in contrast, implies relying on behavioral data assessing the degree of overlap between ground-truth classes and speakers' choice. Though interesting, such evaluation is less crucial given the discrete, non-overlapping nature of the classes in tasks (a) and (c).

The tasks are explored by means of a {MTL} network that jointly performs the three quantification operations (see~\cref{sec:mtl}). The intuition is that solving the lower-level tasks would be beneficial for tackling the higher-level ones. In particular, providing a proportional estimation (`$80$\%') after performing vagueQ (`most') and setComp (`more') should lead to a higher accuracy in the highest-level task, which represents a further step in complexity compared to the previous ones. Moreover, lower-level tasks might be boosted in accuracy by the higher-level ones, since the latter include all the operations that are needed to carry out the former. In addition to the {MTL} model, we test a number of `one-task' networks specifically designed to solve one task at a time (see~\cref{sec:onetaksm}).

\subsection{Dataset}\label{sec:datap}

\begin{figure}[t!]
\hfill
\subfigure{\includegraphics[width=0.485\columnwidth]{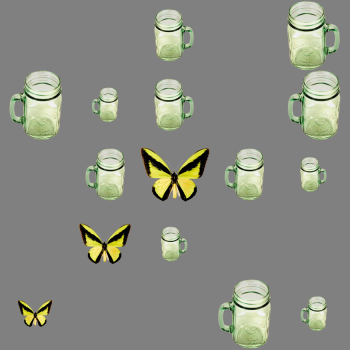}}
\hfill
\subfigure{\includegraphics[width=0.485\columnwidth]{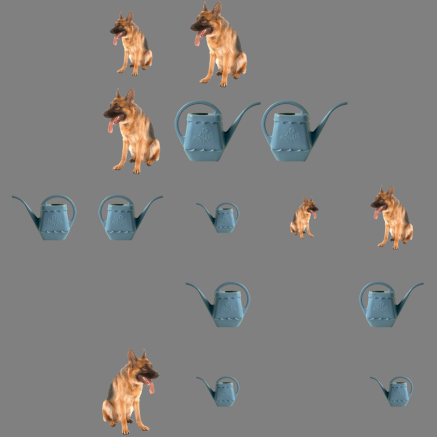}}
\hfill
\caption{Two scenes included in our dataset. The letfmost one depicts a ratio $1$:$4$ ($3$ animals, $12$ artifacts, $15$ total items), the rightmost one a ratio $2$:$3$ ($6$, $9$, $15$).}\label{fig:data}
\end{figure}

We built a large dataset of synthetic visual scenes depicting a variable number of animals and artifacts on the top of a neutral, grey background (see Figure~\ref{fig:data}). In doing so, we employed the same methodology and materials used in~\citet{anonymous}, where the use of quantifiers in \emph{grounded} contexts was explored by asking participants to select the most suitable quantifier for a given scene. Since the category of animals was always treated as the `target', and that of artifacts as the `non-target', we will henceforth use this terminology throughout the paper. The scenes were automatically generated by an in-house script using the following pipeline: (a) Two natural images, one depicting a target object (e.g. a butterfly) and one depicting a non-target (e.g. a mug) were randomly picked up from a sample of the dataset by~\citet{kiani2007}. The sample was obtained by~\citet{anonymous}, who manually selected pictures depicting whole items (not just parts) and whose color, orientation and shape were not deceptive. In total, 100 unique instances of animals and 145 unique instances of artifacts were included; (b) The proportion of targets in the scene (e.g. $20$\%) was chosen by selecting one among $17$ pre-defined \textit{ratios} between targets:non-targets (e.g. $1$:$4$, `four non-targets to one target'). Out of $17$ ratios, $8$ were positive (targets $>$ $50$\%), $8$ negative (targets $<$ $50$\%), and $1$ equal (targets = $50$\%); (c) The absolute number of targets/non-targets was chosen to equally represent the various combinations available for a given ratio (e.g., for ratio $1$:$4$: $1$-$4$, $2$-$8$, $3$-$12$, $4$-$16$), with the constraint of having a number of total objects in the scene (targets+non-targets) ranging from $3$ to $20$. In total, $97$ combinations were represented in the dataset, with an average of $5.7$ combinations/ratio (min $2$, max $18$); (d) To inject some variability, the instances of target/non-target objects were randomly resized according to one of three possible sizes (i.e. medium, big, and small) and flipped on the vertical axis before being randomly inserted onto a $5$*$5$-cell virtual grid. As reported in Table~\ref{tab:stats}, $17$K scenes balanced per ratio ($1$K scenes/ratio) were generated and further split into train ($70$\%), validation ($10$\%), and test ($20$\%).

\begin{table}[t!]
\centering
\small
\begin{tabular}{c|c|c|c|c|}
\cline{2-5}
                                    & train & val & test & \textbf{total} \\ \hline
\multicolumn{1}{|c|}{no. datapoints} & 11.9K & 1.7K       & 3.4K & \textbf{17K}   \\ \hline
\multicolumn{1}{|c|}{\% datapoints} & 70\% & 10\%       & 20\% & \textbf{100\%} \\ \hline
\end{tabular}
\caption{Number and partitioning of the datapoints.}\label{tab:stats}
\end{table}

Ground-truth classes for the tasks of setComp and propTarg were automatically assigned to each scene while generating the data. For vagueQ, we took the probability distributions obtained on a dataset of $340$ scenes by~\citet{anonymous} and we applied them to our datapoints, which were built in the exact same way. These probability distributions had been collected by asking participants to select, from a list of $9$ quantifiers (reported in~\cref{sec:tasks}), the most suitable one to describe the target objects in a visual scene presented for $1$ second. In particular, they were computed against the proportion of targets in the scene, which in that study was shown to be the overall best predictor for quantifiers. To illustrate, given a scene containing $20\%$ of targets (cf. leftmost panel in Figure~\ref{fig:data}), the probability of choosing `few' (ranging from $0$ to $1$) is $0.38$, `almost none' $0.27$, `the smaller part' $0.25$, etc. It is worth mentioning that, for scenes containing either $100$\% or $0$\% targets the probability of choosing `all' and `none', respectively, is around $1$. In all other cases, the distribution of probabilities is fuzzier and reflects the largely overlapping use of quantifiers, as in the example above. On average, the probability of the most-chosen quantifier across ratios is $0.53$. Though this number cannot be seen as a genuine inter-annotator agreement score, it suggests that, on average, there is one quantifier which is preferred over the others.

\begin{figure*}[t!]
\includegraphics[width=\textwidth]{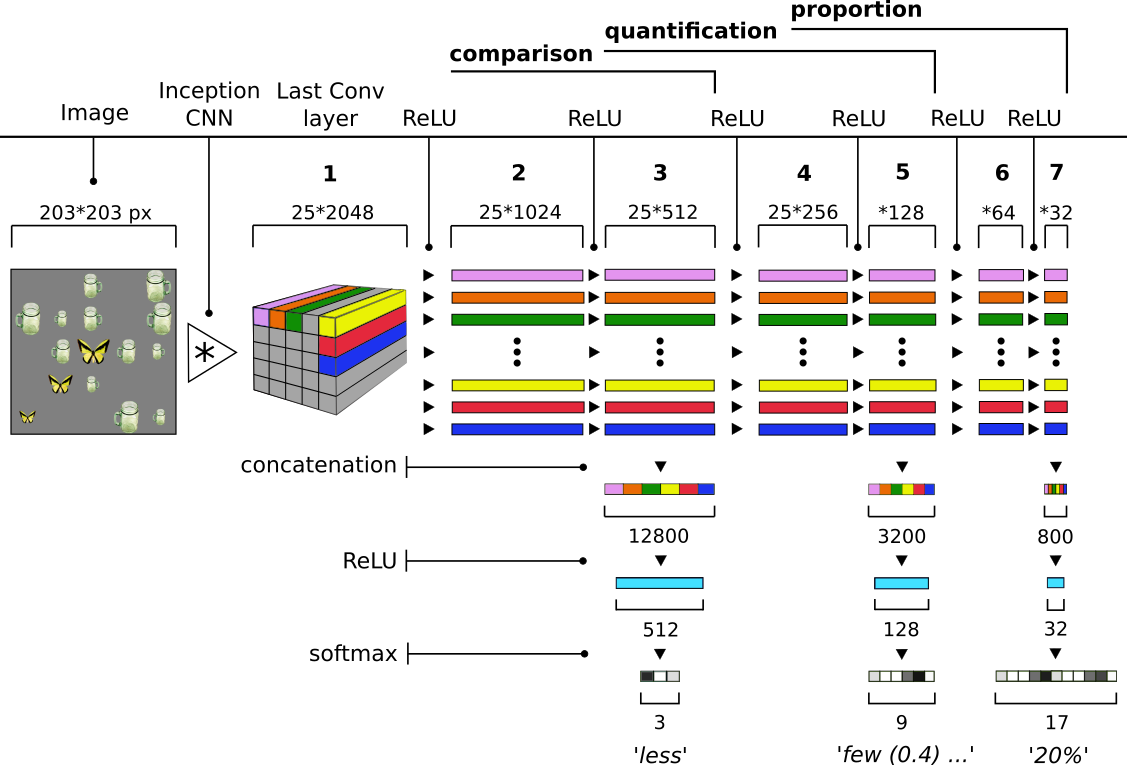}
\caption{Architecture of the \texttt{multi-task-prop} model jointly performing (a) set comparison, (b) vague quantification, and (c) proportional estimation. Given a $203$*$203$-pixel image as input, the model extracts a $25$*$2048$ representation from the last Convolutional layer of the Inception v3. Subsequently, the vectors are reduced twice via ReLU hidden layers to $1024$ and $512$ dimensions. The $512$-d vectors are concatenated and reduced, then a softmax layer is applied to output a $3$-d vector with probability distributions for task (a). The same structure (i.e., $2$ hidden layers, concatenation, reduction, and softmax) is repeated for tasks (b) and (c). All the tasks are trained with cross-entropy. To evaluate tasks (a) and (c), in testing, we extract the highest-probability class and compute \textbf{accuracy}, whereas task (b) is evaluated via \textbf{Pearson's correlation} against the $9$-d ground-truth probability vector.}\label{fig:model}
\end{figure*}

\section{Models}
\label{sec:experiments}

In this section, we describe the various models implemented to perform the tasks. For each model, several settings and parameters were evaluated by means of a thorough ablation analysis. Based on a number of factors like performance, speed, and stability of the networks, we opted for using ReLU nonlinear activation at all hidden layers and the simple and effective Stochastic Gradient Descent ({SGD}) as optimizer (lr = $0.01$). We run each model for $100$ epochs and saved weights and parameters of the epoch with the lowest validation loss. The best model was then used to obtain the predictions in the test set. All models were implemented using Keras.\footnote{\texttt{https://keras.io/}}

\subsection{One-Task Models}\label{sec:onetaksm}

We implemented separate models to tackle one task at a time. For each task, in particular, both a network using `frozen' (i.e. pretrained) visual features and one computing the visual features in an `end-to-end' fashion were tested.
\paragraph{One-Task-Frozen}

These models are simple, $2$-layer (ReLU) Multi-Layer Perceptron ({MLP}) networks that take as input a $2048$-d frozen representation of the scene and output a vector containing softmax probability values. The frozen representation of the scene had been previously extracted using the state-of-art Inception v3 CNN~\cite{inception} pretrained on ImageNet~\cite{imagenet}. In particular, the network is fed with the average of the features computed by the last Convolutional layer, which has size $25$*$2048$.

\paragraph{One-Task-End2end}

These models are {MLP} networks that take as input the $203$*$203$-pixel image and compute the visual features by means of the embedded Inception v3 module, which outputs $25$*$2048$-d vectors (the grey and colored box in Figure~\ref{fig:toymodel}). Subsequently, the $25$ feature vectors are reduced twice via ReLU hidden layers, then concatenated, reduced (ReLU), and fed into a softmax layer to obtain the probability values.

\subsection{Multi-Task Model}\label{sec:mtl}

The \texttt{multi-task-prop} model performs $3$ tasks at the \textit{same time} with an architecture that reproduces in its \textit{order} the conjectured complexity (see Figure~\ref{fig:model} and its caption for technical details). The model has a core structure, represented by layers $1$-$5$ in the figure, which is \textit{shared} across tasks and trained with multiple outputs. In particular, (a) layers $1$, $2$, and $3$ are trained using information regarding the output of all $3$ tasks. That is, these layers are updated three times by as many backpropagation passes: One on the top of setComp output, the second on the top of vagueQ output, the third on the top of propTarg output; (b) layers $4$ and $5$ are affected by information regarding the output of vagueQ and propTarg, and thus updated twice; (c) layers $6$ and $7$ are updated once, on the top of the output of propTarg. Importantly, the three lower layers in Figure~\ref{fig:model} (concatenation, ReLU, softmax) are not shared between the tasks, but specialized to output each a specific prediction. As can be noted, the order of the tasks reflects their complexity, since the last task in the pipeline has $2$ more layers than the preceding one and $4$ more than the first one.

\section{Results}
\label{sec:results}

Table~\ref{tab:acc} reports the performance of each model in the various tasks (note that the lowest row and the rightmost column report results described in~\cref{sec:numbloop}). In setComp, all the models are neatly above chance/majority level ($0.47$). In particular, the \texttt{one-task-end2end} model achieves a remarkable $0.90$ acc., which is more than $10$\% better compared to the simple \texttt{one-task-frozen} model ($0.78$). The same pattern of results can be observed for vagueQ, where the Pearson's correlation (\textit{r}) between the ground-truth and the predicted probability vector is around $0.96$, that is more than $30$\% over the simpler model ($0.62$). This gap increases even more in propTarg, where the accuracy of the frozen model is more than $40$ points below the one achieved by the \texttt{one-task-end2end} model ($0.21$ against $0.66$). These results firmly indicate that, on the one hand, the frozen representation of the visual scene encodes little information about the proportion of targets (likely due to the the different task for which they were pretrained, i.e. object classification). On the other hand, computing the visual features in an end-to-end fashion leads to a significant improvement, suggesting that the network learns to pay attention to features that are helpful for specific tasks.

\begin{table*}[t!]
\centering
\small
\begin{tabular}{|c|c|c|c| |c|}
\hline
\textbf{model}                          & \textbf{setComp}  & \textbf{vagueQ}  & \textbf{propTarg} & \textbf{nTarg}    \\ \hline
                                        & \textit{accuracy} & \textit{Pearson r} & \textit{accuracy} & \textit{accuracy} \\ \hline
\textit{chance/majority}                & 0.470             & 0.320            & 0.058             & 0.132             \\ \hline
one-task-frozen                         & 0.783             & 0.622            & 0.210             & 0.312             \\ \hline
one-task-end2end                        & 0.902             & 0.964            & 0.659             & \textbf{0.966}    \\ \hline
multi-task-prop                         & \textbf{0.995}    & \textbf{0.982}   & \textbf{0.918}    & --                \\ \hline \hline
\multicolumn{1}{|l|}{multi-task-number} & 0.854             & 0.807            & --                & 0.478             \\ \hline
\end{tabular}
\caption{Performance of the models in the tasks of set comparison (setComp), vague quantification (vagueQ), proportional estimation (propTarg), and absolute number of targets (nTarg). Values in \textbf{bold} are the highest.}\label{tab:acc}
\end{table*}

\begin{figure}[b!]
\includegraphics[width=\columnwidth]{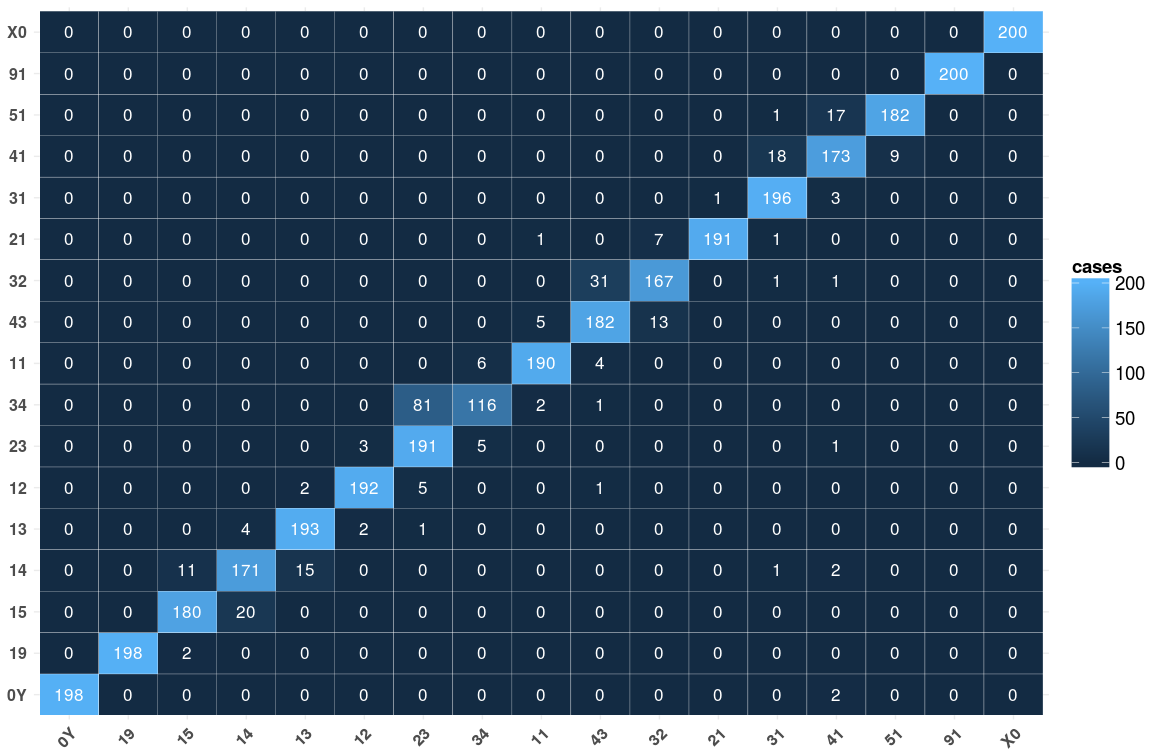}
\caption{PropTarg. Heatmap reporting the errors made by the \texttt{multi-task-prop} model. Note that labels refer to \textit{ratios}, i.e. $14$ stands for ratio $1$:$4$ ($20$\% targets).}\label{fig:ratio-cm}
\end{figure}

The most interesting results, however, are those achieved by the multi-task model, which turns out to be the best in all the tasks. As reported in Table~\ref{tab:acc}, sharing the weights between the various tasks is especially beneficial for propTarg, where the accuracy reaches $0.92$, that is, more than $25$ points over the end-to-end, one-task model. An almost perfect performance of the model in this task can be observed in Figure~\ref{fig:ratio-cm}, which reports the confusion matrix with the errors made by the model. As can be seen, the few errors are between `touching' classes, e.g. between ratio $3$:$4$ ($43$\% of targets) and ratio $2$:$3$ ($40$\%). Since these classes differ by a very small percentage, we gain indirect evidence that the model is learning some kind of proportional information rather than trivial associations between scenes and orthogonal classes.

\begin{figure}[b!]
\includegraphics[width=0.875\columnwidth]{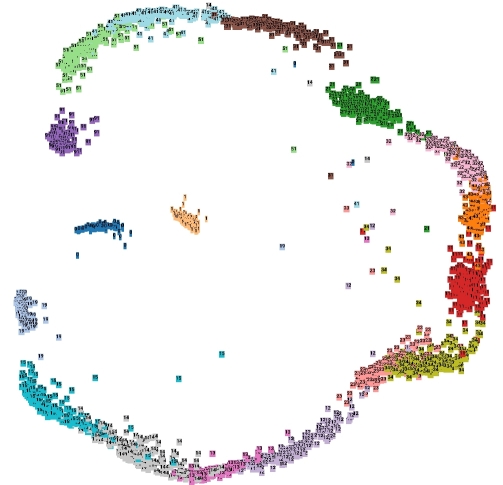}
\caption{PCA visualization of the last layer (before softmax) of the proportional task in the MTL model.}\label{fig:pca}
\end{figure}

To further explore this point, one way is to inspect the last layer of the proportional task (i.e. the 32-d turquoise vector in Figure~\ref{fig:model}). If the vectors contain information regarding the proportion of targets, we should expect scenes depicting the same proportion to have a similar representation. Also, scenes with similar proportions (e.g. $40$\% and $43$\%) would be closer to each other than are scenes with different proportions (e.g. $25$\% and $75$\%). Figure~\ref{fig:pca} depicts the results of a two-dimensional PCA analysis performed on the vectors of the last layer of the proportional task (the $32$-d vectors).\footnote{We used \texttt{https://projector.tensorflow.org/}} As can be noted, scenes depicting the same proportion clearly cluster together, thus indicating that using these representations in a retrieval task would lead to a very high precision. Crucially, the clusters are perfectly ordered with respect to proportion. Starting from the purple cluster on the left side ($90$\%) and proceeding clockwise, we find $83$\% (green), $80$\% (turquoise), $75$\% (brown), and so on, until reaching $10$\% (light blue). Proportions $0$\% (blue) and $100$\% (yellow) are neatly separated from the other clusters, being at the extremes of the `clock'.

\begin{figure}[t!]
\includegraphics[width=\columnwidth]{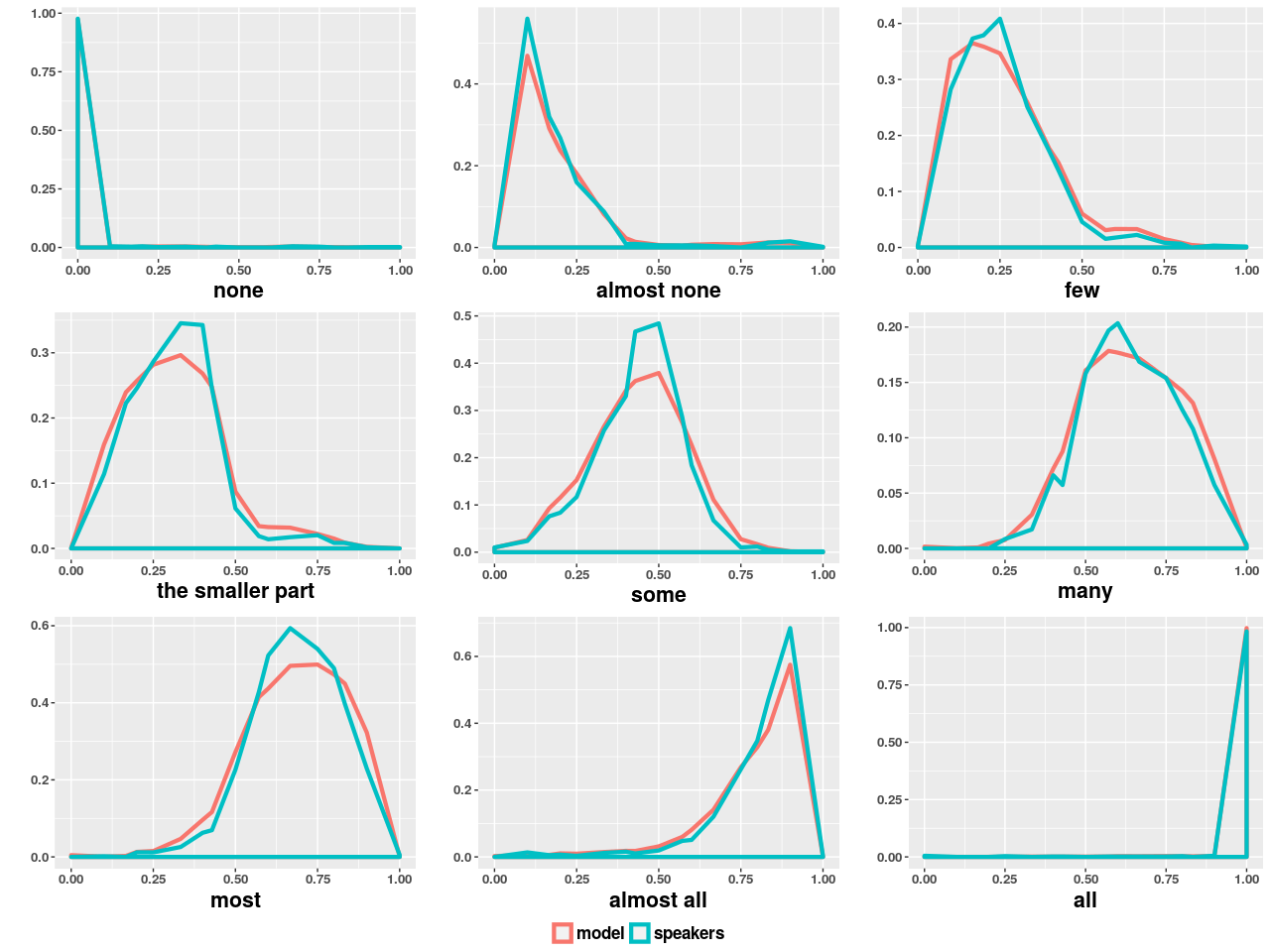}
\caption{VagueQ. Probability values predicted by the \texttt{multi-task-prop} model against ground-truth probability distributions for each quantifier.}\label{fig:probabilities}
\end{figure}

An improvement in the results can be also observed for setComp and vaqueQ, where the model achieves $0.99$ acc. and $0.98$ \textit{r}, respectively. Figure~\ref{fig:probabilities} reports, for each quantifier, the probability values predicted by the model against the ground-truth ones. As can be seen, the red lines (model) approximate very closely the green ones (humans). In the following section, we perform further experiments to provide a deeper evaluation of the results.

\section{In-Depth Evaluation}
\label{sec:further_analysis}

\subsection{Absolute Numbers in the Loop}\label{sec:numbloop}

As discussed in~\cref{sec:introduction}, the cognitive operation underlying setComp, vagueQ, and propTarg is different compared to that of estimating the absolute number of objects included in one set. To investigate whether such dissociation emerges at the computational level, we tested a modified version of our proposed multi-task model where {propTarg} task has been replaced with {nTarg}, namely the task of predicting the absolute number of targets. One-task models were also tested to evaluate the difficulty of the task when performed in isolation. Since the number of targets in the scenes ranges from $0$ to $20$, {nTarg} is evaluated as a $21$-class classification task (majority class $0.13$). 

As reported in Table~\ref{tab:acc}, the accuracy achieved by the \texttt{one-task-end2end} model is extremely high, i.e. around $0.97$. This suggests that, when learned in isolation, the task is fairly easy, but only if the features are computed \textit{within} the model. In fact, using frozen features results in a quite low accuracy, namely $0.31$. This pattern of results is even more interesting if compared against the results of the \texttt{multi-task-number} model. When included in the multi-task pipeline, in fact, nTarg has a huge, $50$-point accuracy drop ($0.48$). Moreover, both setComp and vagueQ turn out to be significantly \textit{hurt} by the highest-level task, and experience a drop of around $14$ and $17$ points compared to the \texttt{one-task-end2end} model, respectively. These findings seem to corroborate the incompatibility of the operations needed for solving the tasks.

\subsection{Reversing the Architecture}

Previous work exploring {MTL} suggested that defining a hierarchy of increasingly-complex tasks is beneficial for jointly learning related tasks (see~\cref{sec:multi}). In the present work, the order of the tasks was inspired by cognitive and linguistic abilities (see~\cref{sec:introduction}). Though cognitively implausible, it might still be the case that the model is able to learn even when reversing the order of the tasks, i.e. from the conjectured highest-level to the lowest-level one. To shed light on this issue, we tested the \texttt{multi-task-prop} model after reversing its architecture. That is, {propTarg} is now the first task, followed by {vagueQ}, and {setComp}.

In contrast with the pattern of results obtained by the original pipeline, no benefits are observed for this version of {MTL} model compared to one-task networks. In particular, both {vagueQ} ($0.32$ \textit{r}) and {propTarg} ($0.08$ acc.) performance are around chance level, with {setComp} reaching just $0.65$ acc., i.e. $25$ point lower than the \texttt{one-task-end2end} model. The pipeline of increasing complexity motivated theoretically is thus confirmed at the computational level.

\subsection{Does {MTL} Generalize?}

\begin{table}[t!]
\centering
\small
\begin{tabular}{|c|c|c|c|}
\hline
\textbf{model}           & \textbf{setComp}  & \textbf{vagueQ}  & \textbf{propTarg} \\ \hline
                         & \textit{accuracy} & \textit{Pearson r} & \textit{accuracy} \\ \hline
\textit{chance/majority} & 0.470             & 0.320            & 0.058             \\ \hline
one-task-frozen          & 0.763             & 0.548            & 0.068             \\ \hline
one-task-end2end         & 0.793             & 0.922            & 0.059             \\ \hline
multi-task-prop          & \textbf{0.943}    & \textbf{0.960}   & \textbf{0.539}    \\ \hline
\end{tabular}
\caption{Unseen dataset. Performance of the models in each task. Values in \textbf{bold} are the highest.}\label{tab:uns}
\end{table}

As discussed in~\cref{sec:multi}, {MTL} is usually claimed to allow a higher generalization power. To investigate whether our proposed \texttt{multi-task-prop} model genuinely learns to quantify from visual scenes, and not just associations between patterns and classes, we tested it with unseen combinations of targets/non-targets. The motivation is that, even in the most challenging {propTarg} task, the model might learn to match a given combination, e.g. $3$:$12$, to a given proportion, i.e. $20$\%. If this is the case, the model would solve the task by learning ``just'' to assign a class to each of the $97$ possible combinations included in the dataset. If it learns a more abstract representation of the proportion of targets depicted in the scene, in contrast, it should be able to generalize to unseen combinations.

\begin{figure}[t!]
\includegraphics[width=\columnwidth]{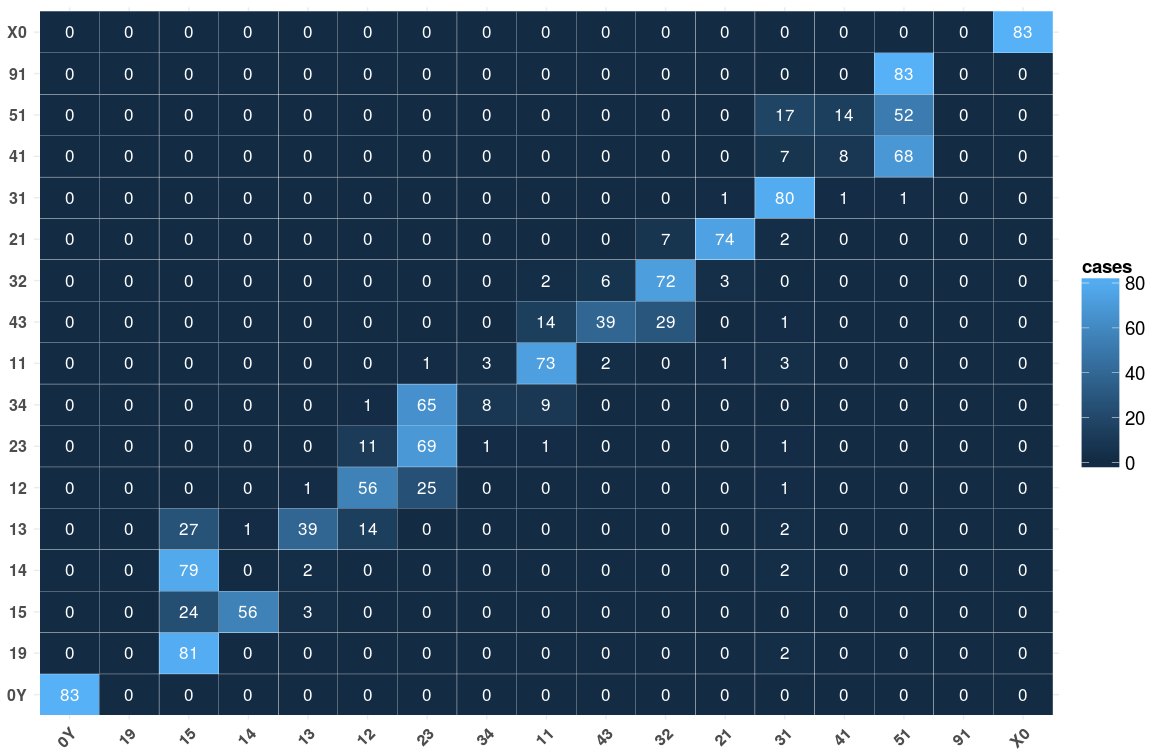}
\caption{PropTarg. Heatmap with the errors made by the \texttt{multi-task-prop} model in the unseen dataset.}\label{fig:uns-ratio-cm}
\end{figure}

We built an additional dataset using the exact same pipeline described in~\cref{sec:datap}. This time, however, we randomly selected one combination per ratio ($17$ combinations in total) to be used only for validation and testing. The remaining $80$ combinations were used for training. A balanced number of datapoints for each combination were generated in val/test, whereas datapoints in training set were balanced with respect to ratios, by randomly selecting scenes among the remaining combinations. The \textit{unseen} dataset included around $14$K datapoints ($80$\% train, $10$\% val, $10$\% test). Table \ref{tab:uns} reports the results of the models on the unseen dataset. Starting from {setComp}, we note a similar and fairly high accuracy achieved by the two one-task models ($0.76$ and $0.79$, respectively). In {vagueQ}, in contrast, the \texttt{one-task-end2end} model neatly outperforms the simpler model ($0.92$ vs. $0.55$ \textit{r}). Finally, in {propTarg} both models are at chance level, with an accuracy that is lower than $0.07$. Overall, this pattern of results suggests that {propTarg} is an extremely hard task for the separate models, which are not able to generalize to unseen combinations. The \texttt{multi-task-prop} model, in contrast, shows a fairly high generalization power. In particular, it achieves $0.54$ acc. in {propTarg}, that is, almost $10$ times chance level. The overall good performance in predicting the correct proportion can be appreciated in Figure~\ref{fig:uns-ratio-cm}, where the errors are represented by means of a heatmap. The error analysis reveals that end-of-the-scale proportions ($0$\% and $100$\%) are the easiest, followed by proportions $75$\% ($3$:$1$), $67$\% ($2$:$1$), $50$\% ($1$:$1$), and $60$\% ($3$:$2$). More in general, negative ratios (targets $<$ $50$\%) are mispredicted to a much greater extent than are positive ones. Moreover, the model shows a bias toward some proportions, that the model seems `to see everywhere'. However, the fact that the errors are found among the adjacent ratios (similar proportions) seems to be a convincing evidence that the model learns representations encoding genuine proportional information. Finally, it is worth mentioning that in {setComp} and {vagueQ} the model achieves very high results, $0.94$ acc. and $0.96$ \textit{r}, respectively.

\section{Discussion}
\label{sec:discussion}

In the present study, we investigated whether \textit{ratio-based} quantification mechanisms, expressed in language by comparatives, quantifiers, and proportions, can be computationally modeled in vision exploiting {MTL}. We proved that sharing a common core turned out to boost the performance in all the tasks, supporting evidence from linguistics, language acquisition, and cognition. Moreover, we showed (a) the increasing complexity of the tasks, (b) the interference of absolute number, and (c) the high generalization power of {MTL}. These results lead to many additional questions. For instance, can these methods be successfully applied to datasets of real scenes? We firmly believe this to be the case, though the results might be affected by the natural biases contained in those images. Also, is this pipeline of increasing complexity specific to vision (non-symbolic level), or is it shared across modalities, \textit{in primis} language? Since linguistic expressions of quantity are grounded on a non-symbolic system, we might expect that a model trained on one modality can be applied to another, at least to some extent. Even further, jointly learning representations from both modalities might represent an even more natural, human-like way to learn and refer to quantities. Further work is needed to explore all these issues.

\section*{Acknowledgments}

We kindly acknowledge Gemma Boleda and the AMORE team (UPF), Raquel Fern{\'a}ndez and the Dialogue Modelling Group (UvA) for the feedback, advice and support. We are also grateful to Aur{\'e}lie Herbelot, Stephan Lee, Manuela Piazza, Sebastian Ruder, and the anonymous reviewers for their valuable comments. This project has received funding from the European Research Council (ERC) under the European Union’s Horizon 2020 research and innovation programme (grant agreement No 715154). We gratefully acknowledge the support of NVIDIA Corporation with the donation of GPUs used for this research. This paper reflects the authors' view only, and the EU is not responsible for any use that may be made of the information it contains.


\begin{flushright}
\includegraphics[width=0.8cm]{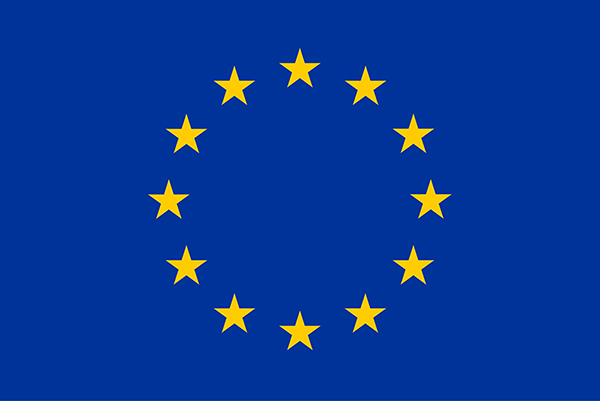}  
\includegraphics[width=0.8cm]{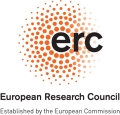} 
\end{flushright}

\newpage

\bibliography{naaclhlt2018}

\newcommand{\noop}[1]{}
\begin{thebibliography}{}
\expandafter\ifx\csname natexlab\endcsname\relax\def\natexlab#1{#1}\fi

\bibitem[{Antol et~al.(2015)Antol, Agrawal, Lu, Mitchell, Batra,
  Lawrence~Zitnick, and Parikh}]{antol2015}
Stanislaw Antol, Aishwarya Agrawal, Jiasen Lu, Margaret Mitchell, Dhruv Batra,
  C~Lawrence~Zitnick, and Devi Parikh. 2015.
\newblock {VQA}: {V}isual {Q}uestion {A}nswering.
\newblock In {\em Proceedings of the IEEE International Conference on Computer
  Vision\/}. pages 2425--2433.

\bibitem[{Arteta et~al.(2016)Arteta, Lempitsky, and Zisserman}]{arteta2016}
Carlos Arteta, Victor Lempitsky, and Andrew Zisserman. 2016.
\newblock Counting in the wild.
\newblock In {\em European Conference on Computer Vision\/}. Springer, pages
  483--498.

\bibitem[{Bingel and S{\o}gaard(2017)}]{bingel2017}
Joachim Bingel and Anders S{\o}gaard. 2017.
\newblock Identifying beneficial task relations for multi-task learning in deep
  neural networks.
\newblock {\em EACL 2017\/} page 164.

\bibitem[{Bryant(2017)}]{bryant2017}
Peter Bryant. 2017.
\newblock {\em Perception and understanding in young children: An experimental
  approach\/}, volume~4.
\newblock Routledge.

\bibitem[{Chattopadhyay et~al.(2017)Chattopadhyay, Vedantam, Selvaraju, Batra,
  and Parikh}]{chatto2016}
Prithvijit Chattopadhyay, Ramakrishna Vedantam, Ramprasaath~R. Selvaraju, Dhruv
  Batra, and Devi Parikh. 2017.
\newblock Counting everyday objects in everyday scenes.
\newblock In {\em The IEEE Conference on Computer Vision and Pattern
  Recognition (CVPR)\/}.

\bibitem[{Collobert et~al.(2011)Collobert, Weston, Bottou, Karlen, Kavukcuoglu,
  and Kuksa}]{collobert2011}
Ronan Collobert, Jason Weston, L{\'e}on Bottou, Michael Karlen, Koray
  Kavukcuoglu, and Pavel Kuksa. 2011.
\newblock Natural language processing (almost) from scratch.
\newblock {\em Journal of Machine Learning Research\/} 12(Aug):2493--2537.

\bibitem[{Deng et~al.(2009)Deng, Dong, Socher, Li, Li, and Fei-Fei}]{imagenet}
Jia Deng, Wei Dong, Richard Socher, Li-Jia Li, Kai Li, and Li~Fei-Fei. 2009.
\newblock {Imagenet: A large-scale hierarchical image database}.
\newblock In {\em Computer Vision and Pattern Recognition, 2009. CVPR 2009.
  IEEE Conference on\/}. IEEE, pages 248--255.

\bibitem[{Fabbri et~al.(2012)Fabbri, Caviola, Tang, Zorzi, and
  Butterworth}]{fabbri2012}
Sara Fabbri, Sara Caviola, Joey Tang, Marco Zorzi, and Brian Butterworth. 2012.
\newblock The role of numerosity in processing nonsymbolic proportions.
\newblock {\em The Quarterly Journal of Experimental Psychology\/}
  65(12):2435--2446.

\bibitem[{Fukui et~al.(2016)Fukui, Park, Yang, Rohrbach, Darrell, and
  Rohrbach}]{fukui2016}
Akira Fukui, Dong~Huk Park, Daylen Yang, Anna Rohrbach, Trevor Darrell, and
  Marcus Rohrbach. 2016.
\newblock Multimodal compact bilinear pooling for visual question answering and
  visual grounding.
\newblock In {\em Conference on Empirical Methods in Natural Language
  Processing\/}. ACL, pages 457--468.

\bibitem[{Girshick(2015)}]{girshick2015}
Ross Girshick. 2015.
\newblock Fast {R}-{CNN}.
\newblock In {\em Proceedings of the IEEE international conference on computer
  vision\/}. pages 1440--1448.

\bibitem[{Halberda and Feigenson(2008)}]{halberda2008b}
Justin Halberda and Lisa Feigenson. 2008.
\newblock {Developmental change in the acuity of the ``Number Sense'': The
  Approximate Number System in 3-, 4-, 5-, and 6-year-olds and adults.}
\newblock {\em Developmental psychology\/} 44(5):1457.

\bibitem[{Halberda et~al.(2008)Halberda, Taing, and Lidz}]{halberda2008}
Justin Halberda, Len Taing, and Jeffrey Lidz. 2008.
\newblock The development of `most' comprehension and its potential dependence
  on counting ability in preschoolers.
\newblock {\em Language Learning and Development\/} 4(2):99--121.

\bibitem[{Hartnett and Gelman(1998)}]{hartnett1998}
Patrice Hartnett and Rochel Gelman. 1998.
\newblock Early understandings of numbers: {P}aths or barriers to the
  construction of new understandings?
\newblock {\em Learning and instruction\/} 8(4):341--374.

\bibitem[{Hashimoto et~al.(2017)Hashimoto, Xiong, Tsuruoka, and
  Socher}]{hashimoto2017}
Kazuma Hashimoto, Caiming Xiong, Yoshimasa Tsuruoka, and Richard Socher. 2017.
\newblock {A Joint Many-Task Model: Growing a Neural Network for Multiple NLP
  Tasks}.
\newblock In {\em Proceedings of the 2017 Conference on Empirical Methods in
  Natural Language Processing (EMNLP)\/}. Association for Computational
  Linguistics, Copenhagen, Denmark, pages 446--456.

\bibitem[{Healey et~al.(1996)Healey, Booth, and Enns}]{healey1996}
Christopher~G Healey, Kellogg~S Booth, and James~T Enns. 1996.
\newblock High-speed visual estimation using preattentive processing.
\newblock {\em ACM Transactions on Computer-Human Interaction (TOCHI)\/}
  3(2):107--135.

\bibitem[{Hurewitz et~al.(2006)Hurewitz, Papafragou, Gleitman, and
  Gelman}]{hurewitz2006}
Felicia Hurewitz, Anna Papafragou, Lila Gleitman, and Rochel Gelman. 2006.
\newblock Asymmetries in the acquisition of numbers and quantifiers.
\newblock {\em Language learning and development\/} 2(2):77--96.

\bibitem[{Johnson et~al.(2017)Johnson, Hariharan, van~der Maaten, Fei-Fei,
  Zitnick, and Girshick}]{clevr2017}
Justin Johnson, Bharath Hariharan, Laurens van~der Maaten, Li~Fei-Fei,
  C~Lawrence Zitnick, and Ross Girshick. 2017.
\newblock {CLEVR}: {A} diagnostic dataset for compositional language and
  elementary visual reasoning.
\newblock In {\em 2017 IEEE Conference on Computer Vision and Pattern
  Recognition (CVPR)\/}. IEEE, pages 1988--1997.

\bibitem[{Kiani et~al.(2007)Kiani, Esteky, Mirpour, and Tanaka}]{kiani2007}
Roozbeh Kiani, Hossein Esteky, Koorosh Mirpour, and Keiji Tanaka. 2007.
\newblock Object category structure in response patterns of neuronal population
  in monkey inferior temporal cortex.
\newblock {\em Journal of neurophysiology\/} 97(6):4296--4309.

\bibitem[{Le~Corre and Carey(2007)}]{le2007}
Mathieu Le~Corre and Susan Carey. 2007.
\newblock One, two, three, four, nothing more: An investigation of the
  conceptual sources of the verbal counting principles.
\newblock {\em Cognition\/} 105(2):395--438.

\bibitem[{Li and Hoiem(2016)}]{li2016}
Zhizhong Li and Derek Hoiem. 2016.
\newblock Learning without forgetting.
\newblock In {\em European Conference on Computer Vision\/}. Springer, pages
  614--629.

\bibitem[{Luong et~al.(2016)Luong, Le, Sutskever, Vinyals, and
  Kaiser}]{luong2016}
Minh-Thang Luong, Quoc~V. Le, Ilya Sutskever, Oriol Vinyals, and Lukasz Kaiser.
  2016.
\newblock Multi-task sequence to sequence learning.
\newblock In {\em International Conference on Learning Representations
  (ICLR)\/}. San Juan, Puerto Rico.

\bibitem[{Malinowski et~al.(2015)Malinowski, Rohrbach, and
  Fritz}]{malinowski2015}
Mateusz Malinowski, Marcus Rohrbach, and Mario Fritz. 2015.
\newblock Ask your neurons: A neural-based approach to answering questions
  about images.
\newblock In {\em Proceedings of the IEEE international conference on computer
  vision\/}. pages 1--9.

\bibitem[{Matthews et~al.(2016)Matthews, Lewis, and Hubbard}]{matthews2016ind}
Percival~G Matthews, Mark~Rose Lewis, and Edward~M Hubbard. 2016.
\newblock Individual differences in nonsymbolic ratio processing predict
  symbolic math performance.
\newblock {\em Psychological science\/} 27(2):191--202.

\bibitem[{McCrink and Wynn(2004)}]{mccrink2004}
Koleen McCrink and Karen Wynn. 2004.
\newblock Large-number addition and subtraction by 9-month-old infants.
\newblock {\em Psychological Science\/} 15(11):776--781.

\bibitem[{Minai(2006)}]{minai2006}
Utako Minai. 2006.
\newblock {\em Everyone knows, therefore every child knows: {A}n investigation
  of logico-semantic competence in child language\/}.
\newblock Ph.D. thesis, University of Maryland.

\bibitem[{Misra et~al.(2016)Misra, Shrivastava, Gupta, and Hebert}]{misra2016}
Ishan Misra, Abhinav Shrivastava, Abhinav Gupta, and Martial Hebert. 2016.
\newblock Cross-stitch networks for multi-task learning.
\newblock In {\em Proceedings of the IEEE Conference on Computer Vision and
  Pattern Recognition\/}. pages 3994--4003.

\bibitem[{Moss and Case(1999)}]{moss1999}
Joan Moss and Robbie Case. 1999.
\newblock Developing children's understanding of the rational numbers: A new
  model and an experimental curriculum.
\newblock {\em Journal for research in mathematics education\/} pages 122--147.

\bibitem[{Odic et~al.(2013)Odic, Pietroski, Hunter, Lidz, and
  Halberda}]{odic2013}
Darko Odic, Paul Pietroski, Tim Hunter, Jeffrey Lidz, and Justin Halberda.
  2013.
\newblock Young children's understanding of ``more'' and discrimination of
  number and surface area.
\newblock {\em Journal of Experimental Psychology: Learning, Memory, and
  Cognition\/} 39(2):451.

\bibitem[{Pezzelle et~al.(\noop{3001}under review)Pezzelle, Bernardi, and
  Piazza}]{anonymous}
Sandro Pezzelle, Raffaella Bernardi, and Manuela Piazza. \noop{3001}under
  review.
\newblock Probing the mental scale of quantifiers.
\newblock {\em Cognition\/} .

\bibitem[{Pezzelle et~al.(2017)Pezzelle, Marelli, and Bernardi}]{pezzelle2017}
Sandro Pezzelle, Marco Marelli, and Raffaella Bernardi. 2017.
\newblock Be precise or fuzzy: Learning the meaning of cardinals and
  quantifiers from vision.
\newblock In {\em Proceedings of the 15th Conference of the European Chapter of
  the Association for Computational Linguistics: Volume 2, Short Papers\/}.
  Association for Computational Linguistics, Valencia, Spain, pages 337--342.

\bibitem[{Piazza(2010)}]{piazza2010}
Manuela Piazza. 2010.
\newblock Neurocognitive start-up tools for symbolic number representations.
\newblock {\em Trends in cognitive sciences\/} 14(12):542--551.

\bibitem[{Piazza and Eger(2016)}]{piazza2016}
Manuela Piazza and Evelyn Eger. 2016.
\newblock Neural foundations and functional specificity of number
  representations.
\newblock {\em Neuropsychologia\/} 83:257--273.

\bibitem[{Ren et~al.(2015)Ren, Kiros, and Zemel}]{ren2015}
Mengye Ren, Ryan Kiros, and Richard Zemel. 2015.
\newblock Exploring models and data for image question answering.
\newblock In {\em Advances in neural information processing systems\/}. pages
  2953--2961.

\bibitem[{Ruder(2017)}]{ruder2017}
Sebastian Ruder. 2017.
\newblock An overview of multi-task learning in deep neural networks.
\newblock {\em arXiv preprint arXiv:1706.05098\/} .

\bibitem[{Segu{\'\i} et~al.(2015)Segu{\'\i}, Pujol, and Vitria}]{segui2015}
Santi Segu{\'\i}, Oriol Pujol, and Jordi Vitria. 2015.
\newblock Learning to count with deep object features.
\newblock In {\em Proceedings of the IEEE Conference on Computer Vision and
  Pattern Recognition Workshops\/}. pages 90--96.

\bibitem[{Shen and Sarkar(2005)}]{shen2005}
Hong Shen and Anoop Sarkar. 2005.
\newblock Voting between multiple data representations for text chunking.
\newblock In {\em Conference of the Canadian Society for Computational Studies
  of Intelligence\/}. Springer, pages 389--400.

\bibitem[{Sidney et~al.(2017)Sidney, Thompson, Matthews, and
  Hubbard}]{sidney2017}
Pooja~G Sidney, Clarissa~A Thompson, Percival~G Matthews, and Edward~M Hubbard.
  2017.
\newblock From continuous magnitudes to symbolic numbers: The centrality of
  ratio.
\newblock {\em Behavioral and Brain Sciences\/} 40.

\bibitem[{Sindagi and Patel(2017)}]{sindagi2017}
Vishwanath~A Sindagi and Vishal~M Patel. 2017.
\newblock {CNN}-{B}ased cascaded multi-task learning of high-level prior and
  density estimation for crowd counting.
\newblock In {\em Advanced Video and Signal Based Surveillance (AVSS), 2017
  14th IEEE International Conference on\/}. IEEE, pages 1--6.

\bibitem[{S{\o}gaard and Goldberg(2016)}]{sogaard2016}
Anders S{\o}gaard and Yoav Goldberg. 2016.
\newblock Deep multi-task learning with low level tasks supervised at lower
  layers.
\newblock In {\em Proceedings of the 54th Annual Meeting of the Association for
  Computational Linguistics\/}. volume~2, pages 231--235.

\bibitem[{Sophian(2000)}]{sophian2000}
Catherine Sophian. 2000.
\newblock Perceptions of proportionality in young children: matching spatial
  ratios.
\newblock {\em Cognition\/} 75(2):145 -- 170.

\bibitem[{Sorodoc et~al.(2016)Sorodoc, Lazaridou, Boleda, Herbelot, Pezzelle,
  and Bernardi}]{sorodoc2016}
Ionut Sorodoc, Angeliki Lazaridou, Gemma Boleda, Aur{\'e}lie Herbelot, Sandro
  Pezzelle, and Raffaella Bernardi. 2016.
\newblock {“Look, some green circles!”: Learning to quantify from images}.
\newblock In {\em Proceedings of the 5th Workshop on Vision and Language\/}.
  pages 75--79.

\bibitem[{Sorodoc et~al.(2018)Sorodoc, Pezzelle, Herbelot, Dimiccoli, and
  Bernardi}]{sorodoc2018}
Ionut Sorodoc, Sandro Pezzelle, Aur{\'e}lie Herbelot, Mariella Dimiccoli, and
  Raffaella Bernardi. 2018.
\newblock {Learning quantification from images: A structured neural
  architecture}.
\newblock {\em Natural Language Engineering\/} page 1–30.

\bibitem[{Spinillo and Bryant(1991)}]{spinillo1991}
Alina~G Spinillo and Peter Bryant. 1991.
\newblock {Children's proportional judgments: The importance of “half”}.
\newblock {\em Child Development\/} 62(3):427--440.

\bibitem[{Stoianov and Zorzi(2012)}]{stoianov2012}
Ivilin Stoianov and Marco Zorzi. 2012.
\newblock Emergence of a `visual number sense' in hierarchical generative
  models.
\newblock {\em Nature neuroscience\/} 15(2):194--196.

\bibitem[{Suhr et~al.(2017)Suhr, Lewis, Yeh, and Artzi}]{suhr2017}
Alane Suhr, Mike Lewis, James Yeh, and Yoav Artzi. 2017.
\newblock A corpus of natural language for visual reasoning.
\newblock In {\em 55th Annual Meeting of the Association for Computational
  Linguistics, ACL\/}.

\bibitem[{Sun et~al.(2017)Sun, Wang, Li, Lv, and Wu}]{sun2017}
Maojin Sun, Yan Wang, Teng Li, Jing Lv, and Jun Wu. 2017.
\newblock Vehicle counting in crowded scenes with multi-channel and multi-task
  convolutional neural networks.
\newblock {\em Journal of Visual Communication and Image Representation\/}
  49:412--419.

\bibitem[{Szegedy et~al.(2016)Szegedy, Vanhoucke, Ioffe, Shlens, and
  Wojna}]{inception}
Christian Szegedy, Vincent Vanhoucke, Sergey Ioffe, Jon Shlens, and Zbigniew
  Wojna. 2016.
\newblock Rethinking the {I}nception {A}rchitecture for {C}omputer {V}ision.
\newblock In {\em Proceedings of the IEEE Conference on Computer Vision and
  Pattern Recognition\/}. pages 2818--2826.

\bibitem[{Treisman(2006)}]{treisman2006}
Anne Treisman. 2006.
\newblock How the deployment of attention determines what we see.
\newblock {\em Visual Cognition\/} 14(4-8):411--443.
\newblock PMID: 17387378.

\bibitem[{Xu and Spelke(2000)}]{xu2000}
Fei Xu and Elizabeth~S Spelke. 2000.
\newblock Large number discrimination in 6-month-old infants.
\newblock {\em Cognition\/} 74(1):B1--B11.

\bibitem[{Yang et~al.(2015)Yang, Hu, Wu, and Yang}]{yang2015}
Ying Yang, Qingfen Hu, Di~Wu, and Shuqi Yang. 2015.
\newblock {Children’s and adults’ automatic processing of proportion in a
  Stroop-like task}.
\newblock {\em International Journal of Behavioral Development\/}
  39(2):97--104.

\bibitem[{Yim et~al.(2015)Yim, Jung, Yoo, Choi, Park, and Kim}]{yim2015}
Junho Yim, Heechul Jung, ByungIn Yoo, Changkyu Choi, Dusik Park, and Junmo Kim.
  2015.
\newblock Rotating your face using multi-task deep neural network.
\newblock In {\em Proceedings of the IEEE Conference on Computer Vision and
  Pattern Recognition\/}. pages 676--684.

\bibitem[{Zhang et~al.(2015{\natexlab{a}})Zhang, Li, Wang, and
  Yang}]{zhang2015}
Cong Zhang, Hongsheng Li, Xiaogang Wang, and Xiaokang Yang. 2015{\natexlab{a}}.
\newblock Cross-scene crowd counting via deep convolutional neural networks.
\newblock In {\em Proceedings of the IEEE Conference on Computer Vision and
  Pattern Recognition\/}. pages 833--841.

\bibitem[{Zhang et~al.(2015{\natexlab{b}})Zhang, Ma, Sameki, Sclaroff, Betke,
  Lin, Shen, Price, and Mech}]{sos2015}
Jianming Zhang, Shugao Ma, Mehrnoosh Sameki, Stan Sclaroff, Margrit Betke, Zhe
  Lin, Xiaohui Shen, Brian Price, and Radomir Mech. 2015{\natexlab{b}}.
\newblock Salient object subitizing.
\newblock In {\em Proceedings of the IEEE Conference on Computer Vision and
  Pattern Recognition\/}. pages 4045--4054.

\bibitem[{Zhang et~al.(2014)Zhang, Luo, Loy, and Tang}]{zhang2014}
Zhanpeng Zhang, Ping Luo, Chen~Change Loy, and Xiaoou Tang. 2014.
\newblock Facial landmark detection by deep multi-task learning.
\newblock In {\em European Conference on Computer Vision\/}. Springer, pages
  94--108.

\end{thebibliography}
\bibliographystyle{acl_natbib}

\end{document}